\documentclass[10pt,twocolumn,letterpaper]{article}
\newcommand{\tabincell}[2]{\begin{tabular}{@{}#1@{}}#2\end{tabular}}
\usepackage{iccv}
\usepackage{times}
\usepackage{epsfig}
\usepackage{graphicx}
\usepackage{amsmath}
\usepackage{amssymb}

\usepackage[breaklinks=true,bookmarks=false]{hyperref}

\iccvfinalcopy 

\def\iccvPaperID{****} 

\ificcvfinal\fi

\begin{document}

\title{Hybrid CNN-Transformer Model For Facial Affect Recognition In the ABAW4 Challenge}

\author{Lingfeng Wang, Haocheng Li, Chunyin Liu\\
University of Electronic Science and Technology of China\\
Chengdu, China\\
{\tt\small \{wanglingfeng,202122011406,202152011703\}@std.uestc.edu.cn}
}

\maketitle
\ificcvfinal\thispagestyle{empty}\fi

\begin{abstract}
This paper describes our submission to the fourth Affective Behavior Analysis (ABAW) competition. We proposed a hybrid CNN-Transformer model for the Multi-Task-Learning (MTL) and Learning from Synthetic Data (LSD) task. Experimental results on validation dataset shows that our method achieves better performance than baseline model, which verifies that the effectiveness of proposed network.
\end{abstract}

\section{Introduction}

Facial affective behavior recognition has become a research hotspot in the field of computer vision sinse it plays an important role in human-computer interaction. Existing research used different approaches to represent human emotions, such as valence-arousal estimation (VA), facial action unit (AU) detection, and facial expression (Expr) classification.Discrete basic expression categories is still the most popular way to represent facial affect, i.e., anger, disgust, fearful, happy, sad and surprised. Artificially generated data could help model to recognise the basic expressions.

In the challenge for Affective Behavior Analysis in-the-wild (ABAW) Competition \cite{kollias2022eccv,kollias2022abaw,kollias2021distribution,kollias2021affect,kollias2020deep,kollias2020va,kollias2019expression,kollias2019deep,kollias2018photorealistic,zafeiriou2017aff,kollias2017recognition}, the organizers collect a large scale in-the-wild database Aff-Wild2 to provide a benchmark for Multi-Task-Learning (MTL) and Learning from Synthetic Data (LSD) tasks respectively. 

In this paper, we describe our approach for the two challenge in the fourth ABAW competition. Firstly, we designed hybrid CNN-Transformer architecture to leverage spatial attention. We build two model using two different pretrained CNN and ensemble the output of the two model. 


\begin{figure*}
\begin{center}
\includegraphics[width=0.9\linewidth]{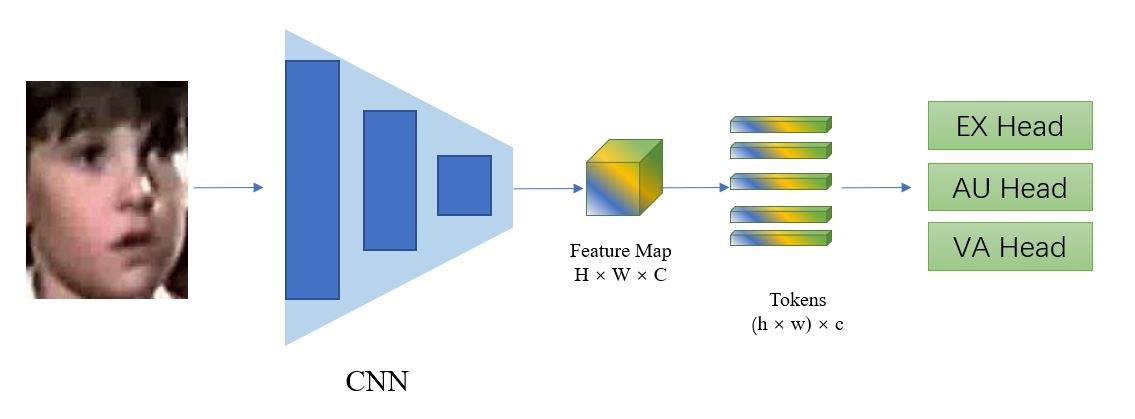}
\end{center}
   \caption{Framework for multi-task affective behavior analysis model.}
\label{fig:model}
\end{figure*}

\section{METHODOLOGY}
\subsection{ Framework}
Figure~\ref{fig:model} shows the framework of our multi-task affective behavior analysis model. The overall architecture of our method is illustrated in Figure 1. Our method is a CNN-Transformer hybrid model which consists of the following two modules: a CNN feature extractor and a Transformer for spatial attention.First, the feature extractor module extracts the local visual features from input images. For feature extraction, we utilized a ResNet-18 network pretrained on AffectNet\cite{affectnet} dataset as well as a HRNe\cite{cheng2020higherhrnet}t pretrained on WFLW \cite{Wu_2018_CVPR}landmark detection dataset.Afterwards, a spatial transformer module consisting of two transformer encoder is used to enhance spatial attention.

\subsection{CNN-Transformer Hybrid Architecture}
We use CNN-Transformer hybrid architecture inspired by \cite{zhao2021former}. CNN-transformer hybrid architecture mainly consists of a ResNet-18 \cite{he2016deep} CNN model and a spatial transformer. The transformer is composed of two transformer encoders described in \cite{vaswani2017attention}. We didn't use pure ViT architecture as \cite{dosovitskiy2020image} due to the fact ViT can capture long-distance feature dependencies effectively but fail to extract local feature details. As for CNN, traditional CNN architecture cannot capture rich global contextual information due to the limit of CNN receptive field. Proposed CNN-transformer hybrid design can leverage both global and local information.

\subsection{Loss Function}
For expression classification task in MTL and LSD challenge, we use cross entropy loss for classification.

Facial AU detection in MTL challenge can be regarded as a multi-label binary classification problem. Weighted BCE allows model to achieve trade-off between recall and precision. The position weights here is proportional to the ratio of positives in the total number for each AU class in training set. 

\begin{equation}
L_{BCE} = \mathbb{E}[-\sum (w_{i}t_{i} \cdot log p_{i} + (1-t_{i})\cdot log(1-p_{i}))] \label{XX}
\end{equation}

The concordance correlation coefficient (CCC) loss \cite{kollias2019expression} is used for valence and arousal estimation in MTL challenge:
\begin{equation}
		L_{V A}=\frac{1}{2} \times(C C C_{V}+C C C_{A})
\end{equation}
\section{ EXPERIMENTAL}
\subsection{Dataset}
Proposed model is fine-tuned on two database for MTL and LSD track respectively. s-Aff-Wild2 database is used for Multi-Task-Learning (MTL) Challenge. It contains selected frames-images from Aff-Wild2 and provide frame-level annotations for valence-arousal estimation, facial action unit detection, and expression classification tasks. We use the official provided cropped images directly. As for LSD task,  the LSD dataset provide 300K synthetic images that contain annotations in terms of the 6 basic facial expressions (anger, disgust, fear, happiness, sadness, surprise). 

\subsection{Training Setup}
ResNet Model is trained on large scale facial expression recognition dataset AffectNet\cite{affectnet}. As for HRNet\cite{cheng2020higherhrnet}, it is trained on facial landmark dataset WFLW\cite{Wu_2018_CVPR}. After that, we freeze the parameters of the CNN and train the spatial transformer on the training set of s-Aff-Wild2 database. Finally, we combine visual branch and audio branch and train joint model. Models are optimized using Adam optimizer and a learning rate of 0.0005. AutoAugment strategy for ImageNet described in \cite{cubuk2019autoaugment} is applied for each input image.  The mini-batch size is set to 64. 

\subsection{Result}
 Result on the validation set of MTL and LSD task is shown in \ref{table:LSDresult}.Our model outperform competition baseline by a lot.
 As for LSD task, both HRNet-Transformer and ResNet-Transformer achieved better performance than competition Baseline. Moreover, if model Ensemble strategy is employed, the F1 score can reach 0.618.
\begin{table}
	\begin{center}
		\begin{tabular}{|l|c|c|c|}
			\hline
			Method & \tabincell{c}{Score MTL}\\
			\hline\hline
			Competition Baseline & 0.3\\
			Ours & 0.981\\
			\hline
		\end{tabular}
	\end{center}
	\caption{performance comparison on validation set for Synthetic Data Challenge}
	\label{table:MTLresult}
\end{table}

\begin{table}
	\begin{center}
		\begin{tabular}{|l|c|c|c|}
			\hline
			Method & \tabincell{c}{Ex (F1)}\\
			\hline\hline
			Competition Baseline & 0.5 \\
			HRNet-Transformer & 0.587\\
			ResNet-Transformer   & 0.596\\
			Ensemble Model &  0.618 \\
			\hline
		\end{tabular}
	\end{center}
	\caption{performance comparison on validation set for Synthetic Data Challenge}
	\label{table:LSDresult}
\end{table}

\section{CONCLUSION}
This paper describe an effective facial action unit detection model by developing a CNN-Transformer hybrid architecture. Our key idea is to firstly use pretrained CNN to extract feature. Then we employ spatial transformer to enhance the relevance of features. Experimental results on validation dataset show that our model outperforms competition baseline, which verifies the effectiveness of proposed method.

{\small
	\bibliographystyle{ieee_fullname}
	\bibliography{egbib}
}
\end{document}